\documentclass[11pt,a4paper]{article}
\usepackage[hyperref]{acl2021}
\usepackage{times}
\usepackage{latexsym}
\usepackage{graphicx}
\usepackage{array}
\usepackage{amsmath}
\usepackage{caption}
\usepackage{subcaption}
\usepackage{amsfonts}
\usepackage{multirow}
\usepackage{url}

\aclfinalcopy 

\newcommand{\vect}[1]{\boldsymbol{#1}}

\title{How is BERT surprised? Layerwise detection of linguistic anomalies}

\author{Bai Li$^{1,4}$, Zining Zhu$^{1,4}$ Guillaume Thomas$^{2}$, Yang Xu$^{1,3,4}$, Frank Rudzicz$^{1,4,5}$ \\
$^1$ University of Toronto, Department of Computer Science \\
$^2$ University of Toronto, Department of Linguistics \\
$^3$ University of Toronto, Cognitive Science Program \\
$^4$ Vector Institute for Artificial Intelligence \;
$^5$ Unity Health Toronto \\
\texttt{\{bai, zining, yangxu, frank\}@cs.toronto.edu} \\ 
\texttt{guillaume.thomas@utoronto.ca}
}

\date{}

\begin{document}
\maketitle
\begin{abstract}

Transformer language models have shown remarkable ability in detecting when a word is anomalous in context, but likelihood scores offer no information about the {\em cause} of the anomaly. In this work, we use Gaussian models for density estimation at intermediate layers of three language models (BERT, RoBERTa, and XLNet), and evaluate our method on BLiMP, a grammaticality judgement benchmark. In lower layers, surprisal is highly correlated to low token frequency, but this correlation diminishes in upper layers. Next, we gather datasets of morphosyntactic, semantic, and commonsense anomalies from psycholinguistic studies; we find that the best performing model RoBERTa exhibits surprisal in earlier layers when the anomaly is morphosyntactic than when it is semantic, while commonsense anomalies do not exhibit surprisal at any intermediate layer. These results suggest that language models employ separate mechanisms to detect different types of linguistic anomalies.

\end{abstract}

\section{Introduction}

\begin{figure}[t]
    \centering
    \includegraphics[width=\linewidth]{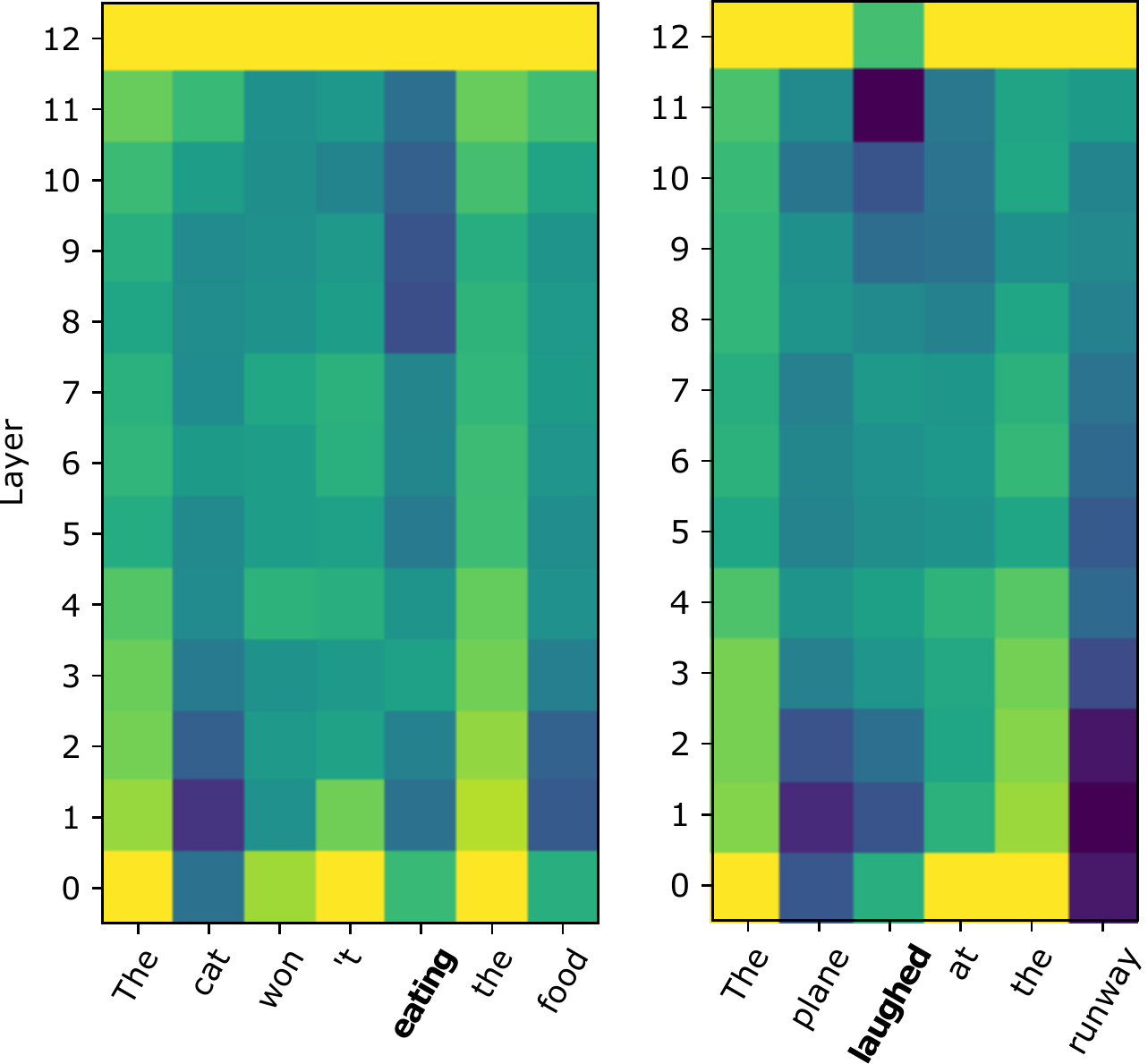}
    \caption{Example sentence with a morphosyntactic anomaly (left) and semantic anomaly (right) (anomalies in bold). Darker colours indicate higher surprisal. We investigate several patterns: first, surprisal at lower layers corresponds to infrequent tokens, but this effect diminishes towards upper layers. Second, morphosyntactic violations begin to trigger high surprisals at an earlier layer than semantic violations.}
    \label{fig:gmm_demo}
\end{figure}

Transformer-based language models (LMs) have achieved remarkable success in numerous natural language processing tasks, prompting many probing studies to determine the extent of their linguistic knowledge. A popular approach is to formulate the problem as a multiple-choice task, where the LM is considered correct if it assigns higher likelihood to the appropriate word than an inappropriate one, given context \citep{gulordava-colorless, ettinger-psycholinguistic, blimp}. The likelihood score, however, only gives a scalar value of the degree that a word is anomalous in context, and cannot distinguish between different {\em ways} that a word might be anomalous.

It has been proposed that there are different types of linguistic anomalies. \citet{chomsky1957} distinguished semantic anomalies ({\em ``colorless green ideas sleep furiously''}) from ungrammaticality ({\em ``furiously sleep ideas green colorless''}). Psycholinguistic studies initially suggested that different event-related potentials (ERPs) are produced in the brain depending on the type of anomaly; e.g., semantic anomalies produce negative ERPs 400 ms after the stimulus, while syntactic anomalies produce positive ERPs 600 ms after  \citep{psycholinguistics-electrified}. Here, we ask whether Transformer LMs show different surprisals in their intermediate layers depending on the type of anomaly. However, LMs do not compute likelihoods at intermediate layers -- only at the final layer.

In this paper, we introduce a new tool to probe for surprisal at intermediate layers of BERT \citep{bert}, RoBERTa \citep{roberta}, and XLNet \citep{xlnet}, formulating the problem as density estimation. We train Gaussian models to fit distributions of embeddings at each layer of the LMs. Using BLiMP \citep{blimp} for evaluation, we show that this model is effective at grammaticality judgement, requiring only a small amount of in-domain text for training. Figure \ref{fig:gmm_demo} shows the method using the RoBERTa model on two example sentences.

We apply our model to test sentences drawn from BLiMP and 7 psycholinguistics studies, exhibiting morphosyntactic, semantic, and commonsense anomalies. We find that morphosyntactic anomalies produce out-of-domain embeddings at earlier layers, semantic anomalies at later layers, and no commonsense anomalies, even though the LM's final accuracy is similar. We show that LMs are internally sensitive to the type of linguistic anomaly, which is not apparent if we only had access to their softmax probability outputs. Our source code and data are available at: \url{https://github.com/SPOClab-ca/layerwise-anomaly}.

\section{Related work}

\subsection{Probing LMs for linguistic knowledge}

Soon after BERT's release, many papers invented probing techniques to discover what linguistic knowledge it contains, and how this information is distributed between layers (e.g., \citet{bertology} provides a comprehensive overview). \citet{bert-rediscovers} used ``edge probing'' to determine each layer's contribution to a task's performance, and discovered that the middle layers contributed more when the task was syntactic, and the upper layers more when the task was semantic.

Several papers found that BERT's middle layers  contain the most syntactic information. \citet{kelly-sentence-probe} found that BERT's middle layers  are best at distinguishing between sentences with direct and indirect object constructions. \citet{hewitt-syntax} used a structural probe to recover syntax trees from contextual embeddings, and found the performance peaked in middle layers.

Probing results are somewhat dependent on the choice of linguistic formalism used to annotate the data, as \citet{ud-vs-sud} found for syntax, and \citet{kuznetsov-framing} found for semantic roles. \citet{miaschi2020linguistic} examined the layerwise performance of BERT for a suite of linguistic features, before and after fine tuning. Our work further investigates what linguistic information is contained in different layers, with a focus on anomalous inputs.

\subsection{Neural grammaticality judgements}

Many recent probing studies used grammaticality judgement tasks to test the knowledge of specific phenomena in LMs. \citet{cola} gathered sentences from linguistic publications, and evaluated by Matthews Correlation with the ground truth. More commonly, the model is presented with a binary choice between an acceptable and unacceptable sentence: BLiMP \citep{blimp} used templates to generate 67k such sentence pairs, covering 12 types of linguistic phenomena. Similarly, \citet{hu-syntax-assessment} created syntactic tests using templates, but defined success criteria using inequalities of LM perplexities.

In contrast with artificial templates, \citet{gulordava-colorless} generated test cases by perturbing natural corpus data to test long-distance dependencies. Most grammaticality studies focused on syntactic phenomena, but \citet{rabinovich-infelicity} tested LMs' sensitivity to semantic infelicities involving indefinite pronouns.

\subsection{Tests of selectional restrictions}

Violations of selectional restrictions are one type of linguistic unacceptability, defined as a semantic mismatch between a verb and an argument. \citet{sasano} examined the geometry of word classes (e.g., words that can be a direct object of the verb `play') in word vector models; they compared single-class models against discriminative models for learning word class boundaries. \citet{chersoni} tested distributional semantic models on their ability to identify selectional restriction violations using stimuli from two psycholinguistic datasets. Finally, \citet{metheniti-selectional}  tested how much BERT relies on selectional restriction information versus other contextual information for making masked word predictions.

\subsection{Psycholinguistic tests for LMs}

The N400 response is a negative event-related potential that occurs roughly 400ms after a stimulus in human brains, and is generally associated with the stimulus being semantically anomalous with respect to the preceding context \citep{kutas-federmeier}. Although many studies have been performed with a diverse range of linguistic stimuli, exactly what conditions trigger the N400 response is still an open question. \citet{frank-erp-surprisal} found that the N400 response is correlated with surprisal, i.e., how unlikely an LM predicts a word given the preceding context.

Recently, several studies have investigated  relationships between surprisal in neural LMs and the N400 response. \citet{michaelov-n400} compared human N400 amplitudes with LSTM-based models using stimuli from several psycholinguistic studies. \citet{ettinger-psycholinguistic} used data from three psycholinguistic studies to probe BERT's knowledge of commonsense and negation. Our work is similar to the latter -- we leverage psycholinguistic studies for their stimuli, but we do not use the their N400 amplitude results.

\begin{figure*}
    \centering
    \begin{subfigure}[t]{0.48\linewidth}
        \includegraphics[width=\linewidth]{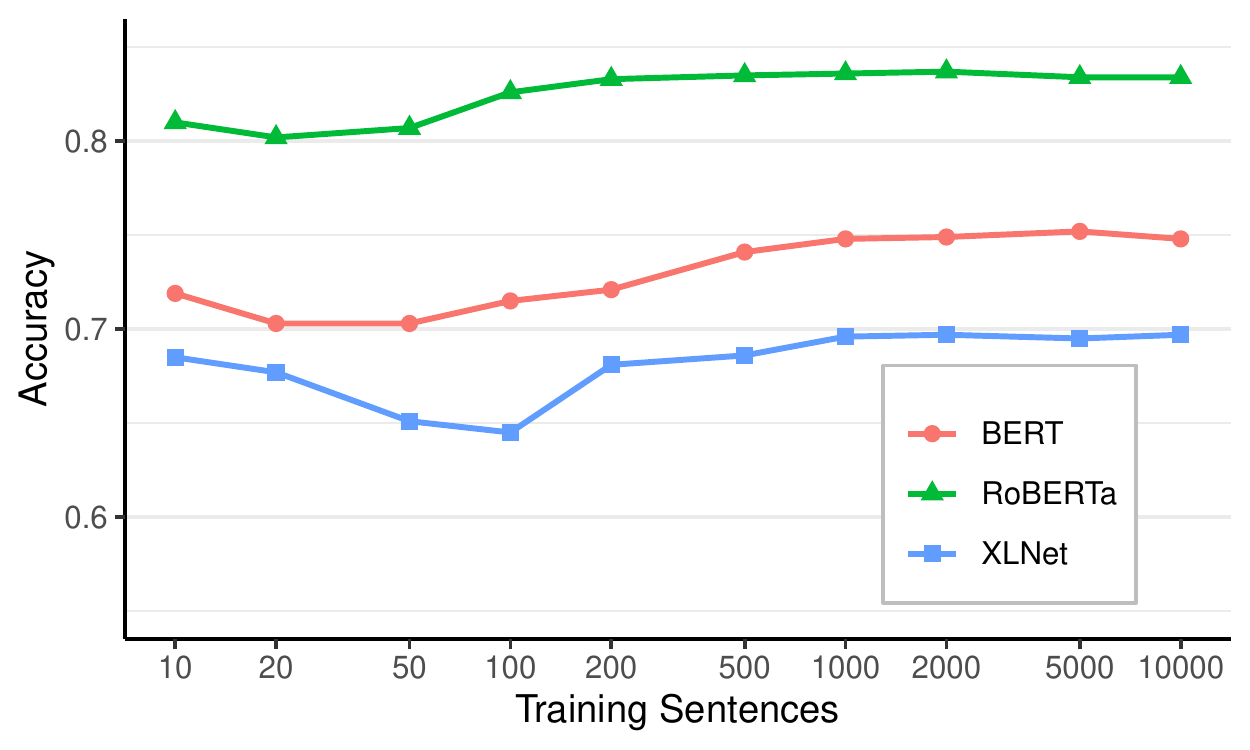}
        \caption{}
        \label{fig:num-sent}
    \end{subfigure}
    \hfill
    \begin{subfigure}[t]{0.48\linewidth}
        \includegraphics[width=\linewidth]{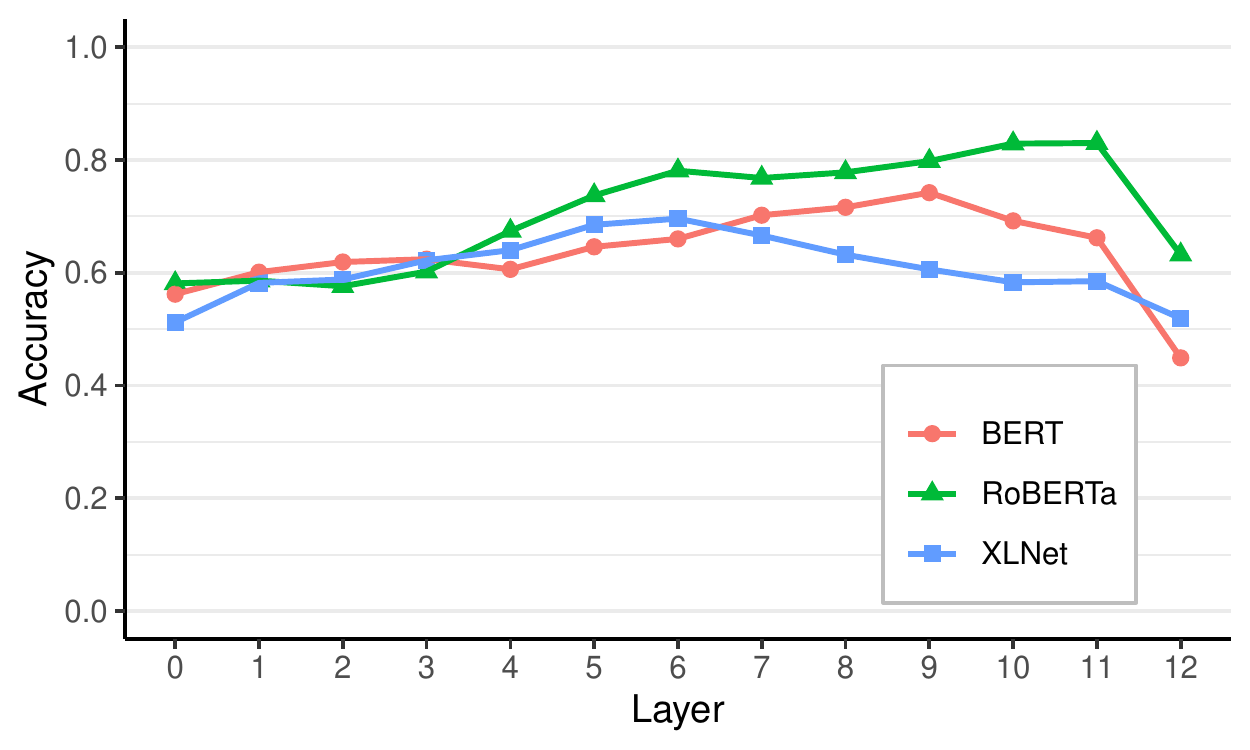}
        \caption{}
        \label{fig:layer-accuracy}
    \end{subfigure}
    \caption{BLiMP accuracy different amounts of training data and across layers, for three LMs. About 1000 sentences are needed before a plateau is reached (mean tokens per sentence = 15.1).}
\end{figure*}

\section{Model}

We use the transformer language model as a contextual embedding extractor (we write this as BERT for convenience). Let $L$ be the layer index, which ranges from 0 to 12 on all of our models. Using a training corpus $\{w_1, \cdots, w_T\}$, we extract contextual embeddings at layer $L$ for each token:
\begin{equation}
    \vect{x}_1^{(L)}, \cdots, \vect{x}_T^{(L)} = \textrm{BERT}_L(w_1, \cdots, w_T).
\end{equation}
Next, we fit a multivariate Gaussian on the extracted embeddings:
\begin{equation}
    \vect{x}_1^{(L)}, \cdots, \vect{x}_T^{(L)} \sim \mathcal{N} (\widehat{\vect{\mu}}_L, \widehat{\vect{\Sigma}}_L).
\end{equation}

For evaluating the layerwise surprisal of a new sentence $\vect{s} = [t_1, \cdots, t_n]$, we similarly extract contextual embeddings using the language model:
\begin{equation}
    \vect{y}_1, \cdots, \vect{y}_n = \textrm{BERT}_L(t_1, \cdots, t_n).
\end{equation}
The surprisal of each token is the negative log likelihood of the contextual vector according to the multivariate Gaussian:
\begin{equation} \label{eq:token-surprisal}
    G_i = -\log p(\vect{y}_i \mid \widehat{\vect{\mu}}_L, \widehat{\vect{\Sigma}}_L) \quad \textrm{for} \ i=1\ldots n.
\end{equation}
Finally, we define the surprisal of sentence $\vect{s}$ as the sum of surprisals of all of its tokens, which is also the joint log likelihood of all of the embeddings:
\begin{multline} \label{eq:sum-agg}
    \textrm{surprisal}_L(t_1, \cdots, t_n) = \sum_{i=1}^n G_i \\
    = -\log p(\vect{y}_1, \cdots, \vect{y}_n \mid \widehat{\vect{\mu}}_L, \widehat{\vect{\Sigma}}_L).
\end{multline}

\subsection{Connection to Mahalanobis distance}

The theoretical motivation for using the sum of log likelihoods is that when we fit a Gaussian model with full covariance matrix, low likelihood corresponds exactly to high Mahalanobis distance from the in-distribution points. The score given by the Gaussian model is:
\begin{multline}
    G = -\log p(\vect{y} \mid \widehat{\vect{\mu}}_L, \widehat{\vect{\Sigma}}_L) \\
    = -\log \left( \frac{1}{(2 \pi)^{D/2} |\widehat{\vect{\Sigma}}_L|^{1/2}} \exp(-\frac{1}{2} d^2) \right),
\end{multline}
where $D$ is the dimension of the vector space, and $d$ is the Mahalanobis distance:
\begin{equation}
    d = \sqrt{(y - \widehat{\vect{\mu}}_L)^T \widehat{\vect{\Sigma}}_L^{-1} (y - \widehat{\vect{\mu}}_L)}.
\end{equation}
Rearranging, we get:
\begin{equation}
    d^2 = 2G -D \log(2 \pi) - \log |\widehat{\vect{\Sigma}}_L|,
\end{equation}
thus the negative log likelihood is the squared Mahalanobis distance plus a constant.

Various methods based on Mahalanobis distance have been used for anomaly detection in neural networks; for example, \citet{lee-anomaly-mahalanobis} proposed a similar method for out-of-domain detection in neural classification models, and \citet{cao-medical-anomaly} found the Mahalanobis distance method to be competitive with more sophisticated methods on medical out-of-domain detection. In Transformer models, \citet{transformer-mahalanobis} used Mahalanobis distance for out-of-domain detection, outperforming methods based on softmax probability and likelihood ratios.

{\bf Gaussian assumptions.} Our model assumes that the embeddings at every layer follow a multivariate Gaussian distribution. Since the Gaussian distribution is the maximum entropy distribution given a mean and covariance matrix, it makes the fewest assumptions and is therefore a reasonable default. \citet{hennigen2020} found that embeddings sometimes do not follow a Gaussian distribution, but it is unclear what alternative distribution would be a better fit, so we will assume a Gaussian distribution in this work.

\subsection{Training and evaluation}

For all of our experiments, we use the `base' versions of pretrained language models BERT \citep{bert}, RoBERTa \citep{roberta}, and XLNet \citep{xlnet}, provided by HuggingFace \citep{huggingface}. Each of these models have 12 contextual layers plus a 0$^\textrm{th}$ static layer, and each layer is 768-dimensional.

We train the Gaussian model on randomly selected sentences from the British National Corpus \citep{bnc}, representative of acceptable English text from various genres. We evaluate on BLiMP \citep{blimp}, a dataset of 67k minimal sentence pairs that test acceptability judgements across a variety of syntactic and semantic phenomena. In our case, a sentence pair is considered correct if the sentence-level surprisal of the unacceptable sentence is higher than that of the acceptable sentence.

{\bf How much training data is needed?} We experiment with training data sizes ranging from 10 to 10,000 sentences (Figure \ref{fig:num-sent}). Compared to the massive amount of data needed for pretraining the LMs, we find that a modest corpus suffices for training the Gaussian anomaly model, and a plateau is reached after 1000 sentences for all three models. Therefore, we use 1000 training sentences (unless otherwise noted) for all subsequent experiments in this paper.

{\bf Which layers are sensitive to anomaly?} We vary $L$ from 0 to 12 in all three models (Figure \ref{fig:layer-accuracy}). The layer with the highest accuracy differs between models: layer 9 has the highest accuracy for BERT, 11 for RoBERTa, and 6 for XLNet. All models experience a sharp drop in the last layer, likely because the last layer is specialized for the MLM pretraining objective.

{\bf Comparisons to other models.} Our best-performing model is RoBERTa, with an accuracy  of 0.830. This is slightly higher the best model reported in BLiMP (GPT-2, with accuracy 0.801). We do not claim to beat the state-of-the-art on BLiMP: \citet{salazar-mlm-scoring} obtains a higher accuracy of 0.865 using RoBERTa-large. Even though the main goal of this paper is not to maximize accuracy on BLiMP, our Gaussian anomaly model is competitive with other transformer-based models on this task.

In Appendix \ref{sec:appendix-ablation}, we explore variations of the Gaussian anomaly model, such as varying the type of covariance matrix, Gaussian mixture models, and one-class SVMs \citep{one-svm}. However, none of these variants offer a significant improvement over a single Gaussian model with full covariance matrix.

\label{sec:best-layer}

\subsection{Lower layers are sensitive to frequency}

We notice that surprisal scores in the lower layers are sensitive to token frequency: higher frequency tokens produce embeddings close to the center of the Gaussian distribution, while lower frequency tokens are at the periphery. The effect gradually diminishes towards the upper layers.

To quantify the sensitivity to frequency, we compute token-level surprisal scores for 5000 sentences from BNC that were not used in training. We then compute the Pearson correlation between the surprisal score and log frequency for each token (Figure \ref{fig:freq-correlation}). In all three models, there is a high correlation between the surprisal score and log frequency at the lower layers, which diminishes at the upper layers. A small positive correlation persists until the last layer, except for XLNet, in which the correlation eventually disappears.

There does not appear to be any reports of this phenomenon in previous work. For static word vectors, \citet{gong-frage} found that embeddings for low-frequency words lie in a different region of the embedding space than high-frequency words. We find evidence that the same phenomenon occurs in contextual embeddings (Appendix \ref{sec:appendix-pca-plots}). In this scenario, the Gaussian model fits the high-frequency region and assigns lower likelihoods to the low-frequency region, explaining the positive correlation at all layers; however, it is still unclear why the correlation diminishes at upper layers.

\section{Levels of linguistic anomalies}

We turn to the question of whether LMs exhibit different behaviour when given inputs with different types of linguistic anomalies. The task of partitioning linguistic anomalies into several distinct classes can be challenging. Syntax and semantics have a high degree of overlap -- there is no widely accepted criterion for distinguishing between ungrammaticality and semantic anomaly (e.g., \citet{abrusan-grammaticality} gives a survey of current proposals), and \citet{poulsen-grammaticality} challenges this dichotomy entirely. Similarly, \citet{warren} noted that semantic anomalies depend somewhat on world knowledge.

Within a class, the anomalies are also heterogeneous (e.g., ungrammaticality may be due to violations of agreement, {\em wh}-movement, negative polarity item licensing, etc), which might each affect the LMs differently. Thus, we define three classes of anomalies that do not attempt to cover all possible linguistic phenomena, but captures different levels of language processing while retaining internal uniformity:

\begin{figure}
    \centering
    \includegraphics[width=\linewidth]{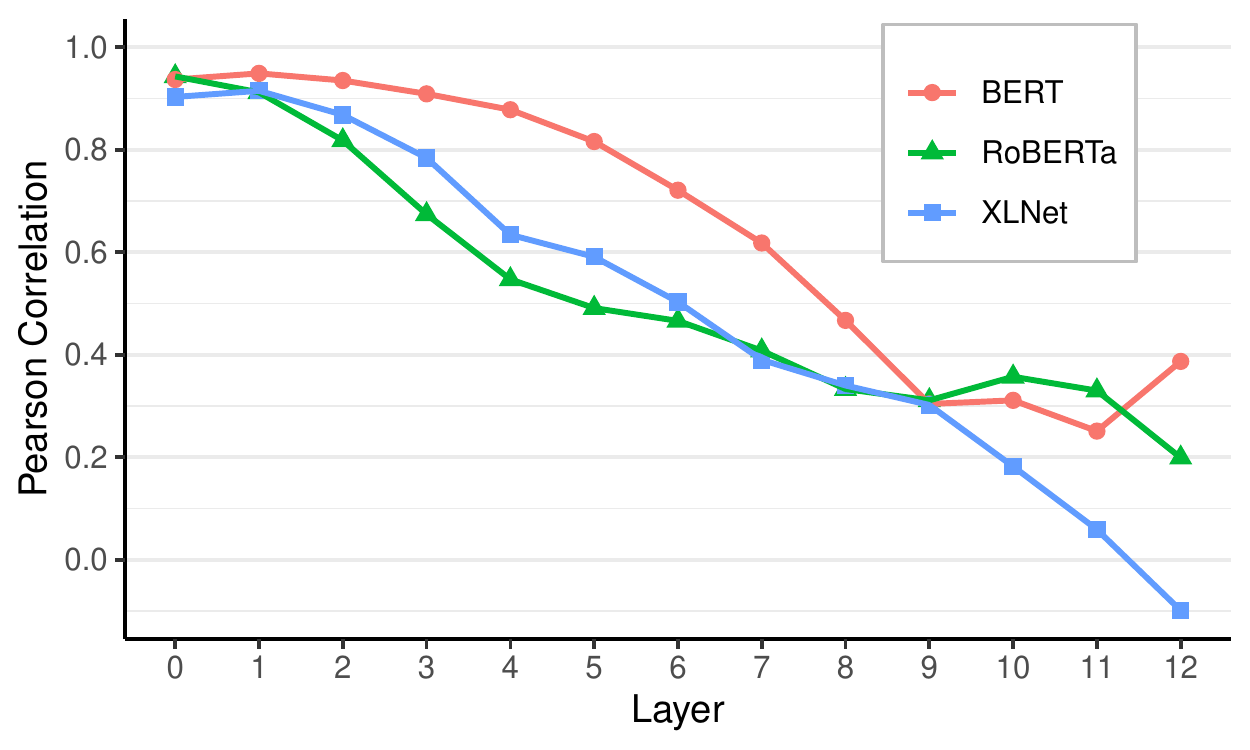}
    \caption{Pearson correlation between token-level surprisal scores (Equation \ref{eq:token-surprisal}) and log frequency. The correlation is highest in the lower layers, and decreases in the upper layers.}
    \label{fig:freq-correlation}
\end{figure}

\begin{enumerate}
    \item {\bf Morphosyntactic anomaly}: an error in the inflected form of a word, for example, subject-verb agreement ({\em \text{*}the boy eat the sandwich}), or incorrect verb tense or aspect inflection ({\em \text{*}the boy eaten the sandwich}). In each case, the sentence can be corrected by changing the inflectional form of one word.
    \item {\bf Semantic anomaly}: a violation of a selectional restriction, such as animacy ({\em \#the house eats the sandwich}). In these cases, the sentence can be corrected by replacing one of the verb's arguments with another one in the same word class that satisfies the verb's selectional restrictions.
    \item {\bf Commonsense anomaly}: sentence describes an situation that is atypical or implausible in the real world but is otherwise acceptable ({\em \#the customer served the waitress}).
\end{enumerate}

\begin{table*}[]
\small
\begin{tabular}{l>{\raggedright}m{0.2\linewidth}ll}
\hline
  \multicolumn{1}{l}{\textbf{Type}} &
  \textbf{Task} &
  \textbf{Correct Example} &
  \textbf{Incorrect Example} \\ \hline
\multirow[b]{3}{*}{\vspace{-2mm}Morphosyntax} &
  BLiMP (Subject-Verb) &
  These casseroles {\bf disgust} Kayla. &
  These casseroles {\bf disgusts} Kayla. \\ [3mm]
&
  BLiMP (Det-Noun) &
  Craig explored that grocery {\bf store}. &
  Craig explored that grocery {\bf stores}. \\ [3mm]
&
  \citet{osterhout-nicol} &
  \begin{tabular}[c]{@{}l@{}}The cats won't {\bf eat} the food that\\ Mary gives them.\end{tabular} &
  \begin{tabular}[c]{@{}l@{}}The cats won't {\bf eating} the food that\\ Mary gives them.\end{tabular} \\ \hline
\multirow[b]{5}{*}{\vspace{-3mm}Semantic} &
  BLiMP (Animacy) &
  \begin{tabular}[c]{@{}l@{}}Amanda was respected by some\\ {\bf waitresses}.\end{tabular} &
  \begin{tabular}[c]{@{}l@{}}Amanda was respected by some\\ {\bf picture}.\end{tabular} \\ [3mm]
&
  \citet{pylkkanen} &
  \begin{tabular}[c]{@{}l@{}}The pilot {\bf flew} the airplane after\\ the intense class.\end{tabular} &
  \begin{tabular}[c]{@{}l@{}}The pilot {\bf amazed} the airplane after\\ the intense class.\end{tabular} \\ [3mm]
&
  \citet{warren} &
  \begin{tabular}[c]{@{}l@{}}Corey's hamster {\bf explored} a nearby\\ backpack and filled it with sawdust.\end{tabular} &
  \begin{tabular}[c]{@{}l@{}}Corey's hamster {\bf entertained} a nearby\\ backpack and filled it with sawdust.\end{tabular} \\ [3mm]
&
  \citet{osterhout-nicol} &
  \begin{tabular}[c]{@{}l@{}}The cats won't {\bf eat} the food that\\ Mary gives them.\end{tabular} &
  \begin{tabular}[c]{@{}l@{}}The cats won't {\bf bake} the food that\\ Mary gives them.\end{tabular} \\ [3mm]
&
  \citet{osterhout-mobley} &
  \begin{tabular}[c]{@{}l@{}}The plane sailed through the air and\\ {\bf landed} on the runway.\end{tabular} &
  \begin{tabular}[c]{@{}l@{}}The plane sailed through the air and\\ {\bf laughed} on the runway.\end{tabular} \\ \hline
\multirow[b]{4}{*}{\vspace{-4mm}Commonsense} &
  \citet{warren} &
  \begin{tabular}[c]{@{}l@{}}Corey's hamster {\bf explored} a nearby\\ backpack and filled it with sawdust.\end{tabular} &
  \begin{tabular}[c]{@{}l@{}}Corey's hamster {\bf lifted} a nearby\\ backpack and filled it with sawdust.\end{tabular} \\ [3mm]
&
  \citet{federmeier-cprag} &
  \begin{tabular}[c]{@{}l@{}}``Checkmate,'' Rosalie announced\\ with glee. She was getting to be\\ really good at {\bf chess}.\end{tabular} &
  \begin{tabular}[c]{@{}l@{}}``Checkmate,'' Rosalie announced\\ with glee. She was getting to be\\ really good at {\bf monopoly}.\end{tabular} \\ [3mm]
&
  \citet{chow-role88} &
  \begin{tabular}[c]{@{}l@{}}The restaurant owner forgot which\\ {\bf customer} the {\bf waitress} had served.\end{tabular} &
  \begin{tabular}[c]{@{}l@{}}The restaurant owner forgot which\\ {\bf waitress} the {\bf customer} had served.\end{tabular} \\ [3mm]
&
  \citet{urbach} &
  \begin{tabular}[c]{@{}l@{}}Prosecutors accuse {\bf defendants} of\\ committing a crime.\end{tabular} &
  \begin{tabular}[c]{@{}l@{}}Prosecutors accuse {\bf sheriffs} of\\ committing a crime.\end{tabular} \\ \hline
\end{tabular}
\caption{Example sentence pair for each of the 12 tasks. The 3 BLiMP tasks are generated from templates; the others are stimuli materials taken from psycholinguistic studies.}
\label{tab:task-example}
\end{table*}

\subsection{Summary of anomaly datasets}

We use two sources of data for experiments on linguistic anomalies: synthetic sentences generated from templates, and materials from psycholinguistic studies. Both have advantages and disadvantages -- synthetic data can be easily generated in large quantities, but the resulting sentences may be odd in unintended ways. Psycholinguistic stimuli are designed to control for confounding factors (e.g., word frequency) and human-validated for acceptability, but are smaller (typically fewer than 100 sentence pairs).

We curate a set of 12 tasks from BLiMP and 7 psycholinguistic studies\footnote{Several of these stimuli have been used in natural language processing research. \citet{chersoni} used the data from \citet{pylkkanen} and \citet{warren} to probe word vectors for knowledge of selectional restrictions. \citet{ettinger-psycholinguistic} used data from \citet{federmeier-cprag} and \citet{chow-role88}, which were referred to as CPRAG-102 and ROLE-88 respectively.}. Each sentence pair consists of a control and an anomalous sentence, so that all sentences within a task differ in a consistent manner. Table \ref{tab:task-example} shows an example sentence pair from each task. We summarize each dataset:

\begin{enumerate}
    \item BLiMP \citep{blimp}: we use subject-verb  and determiner-noun agreement tests as morphosyntactic anomaly tasks. For simplicity, we only use the basic regular sentences, and exclude sentences involving irregular words or distractor items. We also use the two argument structure tests involving animacy as a semantic anomaly task. All three BLiMP tasks therefore have 2000 sentence pairs.
    \item \citet{osterhout-nicol}: contains 90 sentence triplets containing a control, syntactic, and semantic anomaly. Syntactic anomalies involve a modal verb followed by a verb in {\em -ing} form; semantic anomalies have a selectional restriction violation between the subject and verb. There are also double anomalies (simultaneously syntactic and semantic) which we do not use.
    \item \citet{pylkkanen}: contains 70 sentence pairs where the verb is replaced in the anomalous sentence with one that requires an animate object, thus violating the selectional restriction. In half the sentences, the verb is contained in an embedded clause.
    \item \citet{warren}: contains 30 sentence triplets with a possible condition, a selectional restriction violation between the subject and verb, and an impossible condition where the subject cannot carry out the action, i.e., a commonsense anomaly.
    \item \citet{osterhout-mobley}: we use data from experiment 2, containing 90 sentence pairs where the verb in the anomalous sentence is semantically inappropriate. The experiment also tested gender agreement errors, but we do not include these stimuli.
    \item \citet{federmeier-cprag}: contains 34 sentence pairs, where the final noun in each anomalous sentence is an inappropriate completion, but in the same semantic category as the expected completion.
    \item \citet{chow-role88}: contains 44 sentence pairs, where two of the nouns in the anomalous sentence are swapped to reverse their roles. This is the only task in which the sentence pair differs by more than one token.
    \item \citet{urbach}: contains 120 sentence pairs, where the anomalous sentence replaces a patient of the verb with an atypical one.
\end{enumerate}

\subsection{Quantifying layerwise surprisal}

Let $\mathcal{D} = \{(\vect{s}_1, \vect{s}_1'), \cdots, (\vect{s}_n, \vect{s}_n')\}$ be a dataset of sentence pairs, where $\vect{s}_i$ is a control sentence and $\vect{s}_i'$ is an anomalous sentence. For each layer $L$, we define the {\em surprisal gap} as the mean difference of surprisal scores between the control and anomalous sentences, scaled by the standard deviation:
\begin{multline}
    \textrm{surprisal\ gap}_L(\mathcal{D}) = \\ \frac
    {\mathbb{E}\{\textrm{surprisal}_L(\vect{s}'_i) - \textrm{surprisal}_L(\vect{s}_i)\}_{i=1}^n}
    {\sigma\{\textrm{surprisal}_L(\vect{s}'_i) - \textrm{surprisal}_L(\vect{s}_i)\}_{i=1}^n}
\end{multline}

\begin{figure}
    \centering
    \includegraphics[width=0.96\linewidth]{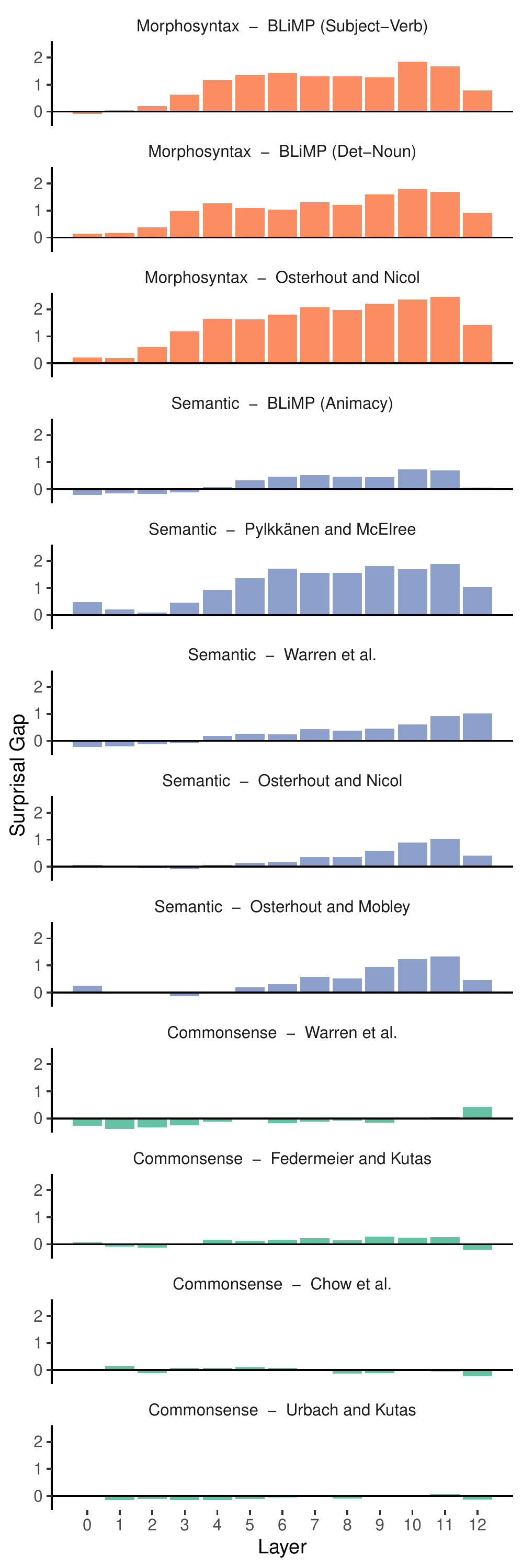}
    \caption{Layerwise surprisal gaps for all tasks using the RoBERTa model. Generally, a positive surprisal gap appears in earlier layers for morphosyntactic tasks than for semantic tasks; no surprisal gap appears at any layer for commonsense tasks.}
    \label{fig:roberta-manual}
\end{figure}

The surprisal gap is a scale-invariant measure of sensitivity to anomaly, similar to a signal-to-noise ratio. While surprisal scores are unitless, the surprisal gap may be viewed as the number of standard deviations that anomalous sentences trigger surprisal above control sentences. This is advantageous over accuracy scores, which treats the sentence pair as correct when the anomalous sentence has higher surprisal by any margin; this hard cutoff masks differences in the magnitude of surprisal. The metric also allows for fair comparison of surprisal scores across datasets of vastly different sizes. Figure \ref{fig:roberta-manual} shows the surprisal gap for all 12 tasks, using the RoBERTa model; the results for BERT and XLNet are in the Appendix \ref{sec:appendix-surprisal-gap}.

Next, we compare the performance of the Gaussian  model with the masked language model (MLM). We score each instance as correct if the masked probability of the correct word is higher than the anomalous word. One limitation of the MLM approach is that it requires the sentence pair to be identical in all places except for one token, since the LMs do not support modeling joint probabilities over multiple tokens. To ensure fair comparison between GM and MLM, we exclude instances where the differing token is out-of-vocabulary in any of the LMs (this excludes approximately 30\% of instances). For the Gaussian  model, we compute accuracy using the best-performing layer for each model (Section \ref{sec:best-layer}). The results are listed in Table \ref{tab:mlm-results}.

\begin{table*}
\centering \small
\begin{tabular}{llrrrrrrr}
\hline
\multirow{2}{*}{\textbf{Type}} &
  \multirow{2}{*}{\textbf{Task}} &
  \multicolumn{1}{c}{\multirow{2}{*}{\textbf{Size}}} &
  \multicolumn{2}{c}{\textbf{BERT}} &
  \multicolumn{2}{c}{\textbf{RoBERTa}} &
  \multicolumn{2}{c}{\textbf{XLNet}} \\
                              &                          & \multicolumn{1}{c}{} & GM          & MLM         & GM          & MLM   & GM          & MLM         \\ \hline
\multirow{3}{*}{Morphosyntax} & BLiMP (Subject-Verb)     & 2000                 & 0.953       & 0.955       & 0.971       & 0.957 & 0.827       & 0.584       \\
                              & BLiMP (Det-Noun)         & 2000                 & 0.970       & 0.999       & 0.983       & 0.999 & 0.894       & 0.591       \\
                              & \citet{osterhout-nicol}  & 90                   & 1.000       & 1.000       & 1.000       & 1.000 & 0.901       & 0.718       \\ \hline
\multirow{5}{*}{Semantic}     & BLiMP (Animacy)          & 2000                 & 0.644       & 0.787       & 0.767       & 0.754 & 0.675       & 0.657       \\
                              & \citet{pylkkanen}        & 70                   & 0.727       & 0.955       & 0.932       & 0.955 & $^{*}$0.636 & 0.727       \\
                              & \citet{warren}           & 30                   & $^{*}$0.556 & 1.000       & 0.944       & 1.000 & $^{*}$0.667 & $^{*}$0.556 \\
                              & \citet{osterhout-nicol}  & 90                   & 0.681       & 0.957       & 0.841       & 1.000 & $^{*}$0.507 & 0.783       \\
                              & \citet{osterhout-mobley} & 90                   & $^{*}$0.528 & 1.000       & 0.906       & 0.981 & $^{*}$0.302 & 0.774       \\ \hline
\multirow{4}{*}{Commonsense} &
  \citet{warren} &
  30 &
  $^{*}$0.600 &
  $^{*}$0.550 &
  0.750 &
  $^{*}$0.450 &
  $^{*}$0.300 &
  $^{*}$0.600 \\
                              & \citet{federmeier-cprag} & 34                   & $^{*}$0.458 & $^{*}$0.708 & $^{*}$0.583 & 0.875 & $^{*}$0.625 & $^{*}$0.667 \\
                              & \citet{chow-role88}      & 44                   & $^{*}$0.591 & n/a         & $^{*}$0.432 & n/a   & $^{*}$0.568 & n/a         \\
                              & \citet{urbach}           & 120                  & $^{*}$0.470 & 0.924       & $^{*}$0.485 & 0.939 & $^{*}$0.500 & 0.712       \\ \hline
\end{tabular}
\caption{Comparing accuracy scores between Gaussian anomaly model (GM) and masked language model (MLM) for all models and tasks. Asterisks indicate that the accuracy is not better than random (0.5), using a binomial test with threshold of $p < 0.05$ for significance. The MLM results for \citet{chow-role88} are excluded because the control and anomalous sentences differ by more than one token. The best layers for each model (Section \ref{sec:best-layer}) are used for GM, and the last layer is used for MLM. Generally, MLM outperforms GM, and the difference is greater for semantic and commonsense tasks.}
\label{tab:mlm-results}
\end{table*}

\section{Discussion}

\subsection{Anomaly type and surprisal}

Morphosyntactic anomalies generally appear earlier than semantic anomalies (Figure \ref{fig:roberta-manual}). The surprisal gap plot exhibits different patterns depending on the type of linguistic anomaly: morphosyntactic anomalies produce high surprisal relatively early (layers 3-4), while semantic anomalies produce low surprisals until later (layers 9 and above). Commonsense anomalies do not result in surprisals at {\em any} layer: the surprisal gap is near zero for all of the commonsense tasks. The observed difference between morphosyntactic and semantic anomalies is consistent with previous work \citep{bert-rediscovers}, which found that syntactic information appeared earlier in BERT than semantic information.

One should be careful and avoid drawing conclusions from only a few experiments. A similar situation occurred in psycholinguistics research \citep{psycholinguistics-electrified}: early results suggested that the N400 was triggered by semantic anomalies, while syntactic anomalies triggered the P600 -- a different type of ERP. However, subsequent experiments found exceptions to this rule, and now it is believed that the N400 cannot be categorized by any standard dichotomy, like syntax versus semantics \citep{kutas-federmeier}. In our case, \citet{pylkkanen} is an exception: the task is a semantic anomaly, but produces surprisals in early layers, similar to the morphosyntactic tasks. Hence it is possible that the dichotomy is something other than syntax versus semantics; we leave to future work to determine more precisely what conditions trigger high surprisals in lower versus upper layers of LMs.

\subsection{Comparing anomaly model with MLM}

The masked language model (MLM) usually outperforms the Gaussian anomaly model (GM), but the difference is uneven. MLM performs much better than GM on commonsense tasks, slightly better on semantic tasks, and about the same or slightly worse on morphosyntactic tasks. It is not obvious why MLM should perform better than GM, but we note two subtle differences between the MLM and GM setups that may be contributing factors. First, the GM method adds up the surprisal scores for the whole sequence, while MLM only considers the softmax distribution at one token. Second, the input sequence for MLM always contains a {\tt [MASK]} token, whereas GM takes the original unmasked sequences as input, so the representations are never identical between the two setups.

MLM generally outperforms GM, but it does not solve every task: all three LMs fail to perform above chance on the data from \citet{warren}. This set of stimuli was designed so that both the control and impossible completions are not very likely or expected, which may have caused the difficulty for the LMs. We excluded the task of \citet{chow-role88} for MLM because the control and anomalous sentences differed by more than one token\footnote{Sentence pairs with multiple differing tokens are inconvenient for MLM to handle, but this is not a fundamental limitation. For example, \citet{salazar-mlm-scoring} proposed a modification to MLM to handle such cases: they compute a {\em pseudo-log-likelihood} score for a sequence by replacing one token at a time with a {\tt [MASK]} token, applying MLM to each masked sequence, and summing up the log likelihood scores.}.

\subsection{Differences between LMs}

RoBERTa is the best-performing of the three LMs in both the GM and MLM settings: this is expected since it is trained with the most data and performs well on many natural language benchmarks. Surprisingly, XLNet is ill-suited for this task and performs worse than BERT, despite having a similar model capacity and training data.

The surprisal gap plots for BERT and XLNet (Appendix \ref{sec:appendix-surprisal-gap}) show some differences from RoBERTa: only morphosyntactic tasks produce out-of-domain embeddings in these two models, and not semantic or commonsense tasks. Evidently, how LMs behave when presented with anomalous inputs is dependent on model architecture and training data size; we leave exploration of this phenomenon to future work.

\section{Conclusion}

We use Gaussian models to characterize out-of-domain embeddings at intermediate layers of Transformer language models. The model requires a relatively small amount of in-domain data. Our experiments reveal that out-of-domain points in lower layers correspond to low-frequency tokens, while grammatically anomalous inputs are out-of-domain in higher layers. Furthermore, morphosyntactic anomalies are recognized as out-of-domain starting from lower layers compared to syntactic anomalies. Commonsense anomalies do not generate out-of-domain embeddings at any layer, even when the LM has a preference for the correct cloze completion. These results show that depending on the type of linguistic anomaly, LMs use different mechanisms to produce the output softmax distribution.

\section*{Acknowledgements}

We thank Julian Salazar and our anonymous reviewers for their helpful suggestions. YX is funded through an NSERC Discovery Grant, a SSHRC Insight Grant, and an Ontario ERA award. FR is supported by a CIFAR Chair in Artificial Intelligence.

\bibliography{acl2021}
\bibliographystyle{acl_natbib}

\clearpage
\appendix

\section{Ablation experiments on Gaussian model}
\label{sec:appendix-ablation}

We compare some variations to our methodology of training the Gaussian model. All of these variations are evaluated on the full BLiMP dataset. In each experiment,  (unless otherwise noted)  the language model is RoBERTa-base, using the second-to-last layer, and the Gaussian model has a full covariance matrix trained with 1000 sentences from the BNC corpus.

{\bf Covariance matrix}. We vary the type of covariance matrix (Table \ref{tab:cov-acc}). Diagonal and spherical covariance matrices perform worse than with the full covariance matrix; this may be expected, as the full matrix has the most trainable parameters.

\begin{table}[h]
\centering
\begin{tabular}{lr}
\hline
\textbf{Covariance} & \multicolumn{1}{l}{\textbf{Accuracy}} \\ \hline
Full              & 0.830                                 \\
Diagonal              & 0.755                                 \\
Spherical         & 0.752                                 \\ \hline
\end{tabular}
\caption{Varying the type of covariance matrix in the Gaussian model.}
\label{tab:cov-acc}
\end{table}

{\bf Gaussian mixture models}. We try GMMs with up to 16 mixture components (Table \ref{tab:gmm-acc}). We observe a small increase in accuracy compared to a single Gaussian, but the difference is too small to justify the increased training time.

\begin{table}[h]
\centering
\begin{tabular}{rr}
\hline
\multicolumn{1}{l}{\textbf{Components}} & \multicolumn{1}{l}{\textbf{Accuracy}} \\ \hline
1                                       & 0.830                                 \\
2                                       & 0.841                                 \\
4                                       & 0.836                                 \\
8                                       & 0.849                                 \\
16                                      & 0.827                                 \\ \hline
\end{tabular}
\caption{Using Gaussian mixture models (GMMs) with multiple components.}
\label{tab:gmm-acc}
\end{table}

{\bf Genre of training text}. We sample from genres of BNC (each time with 1000 sentences) to train the Gaussian model (Table \ref{tab:genre-acc}). The model performed worse when trained with the academic and spoken genres, and about the same with the fiction and news genres, perhaps because their vocabularies and grammars are more similar to those in the BLiMP sentences.

\begin{table}[h]
\centering
\begin{tabular}{lr}
\hline
\textbf{Genre} & \multicolumn{1}{l}{\textbf{Accuracy}} \\ \hline
Academic            & 0.797                                 \\
Fiction            & 0.840                                 \\
News           & 0.828                                 \\
Spoken            & 0.795                                 \\
All            & 0.830                                 \\ \hline
\end{tabular}
\caption{Effect of the genre of training data.}
\label{tab:genre-acc}
\end{table}

{\bf One-class SVM}. We try replacing the Gaussian model with a one-class SVM \citep{one-svm}, another popular model for anomaly detection. We use the default settings from scikit-learn with three kernels (Table \ref{tab:1svm-acc}), but it performs worse than the Gaussian model on all settings.

\begin{table}[h]
\centering
\begin{tabular}{lr}
\hline
\textbf{Kernel} & \multicolumn{1}{l}{\textbf{Score}} \\ \hline
RBF             & 0.738                              \\
Linear          & 0.726                              \\
Polynomial            & 0.725                              \\ \hline
\end{tabular}
\caption{Using 1-SVM instead of GMM, with various kernels.}
\label{tab:1svm-acc}
\end{table}

{\bf Sentence aggregation}. Instead of Equation \ref{eq:sum-agg}, we try defining sentence-level surprisal as the maximum surprisal among all tokens (Table \ref{tab:max-acc}):
\begin{equation}
    \textrm{surprisal}(s_1, \cdots, s_n) = \textrm{max}_{i=1}^n G_i; \\
\end{equation}
however, this performs worse than using the sum of token surprisals.

\begin{table}[h]
\centering
\begin{tabular}{lr}
\hline
\textbf{Aggregation} & \multicolumn{1}{l}{\textbf{Accuracy}} \\ \hline
Sum                  & 0.830                              \\
Max                  & 0.773                              \\ \hline
\end{tabular}
\caption{Two sentence-level aggregation strategies}
\label{tab:max-acc}
\end{table}

\newpage
\section{PCA plots of infrequent tokens}
\label{sec:appendix-pca-plots}

\begin{figure}
    \centering
    \includegraphics[width=0.7\linewidth]{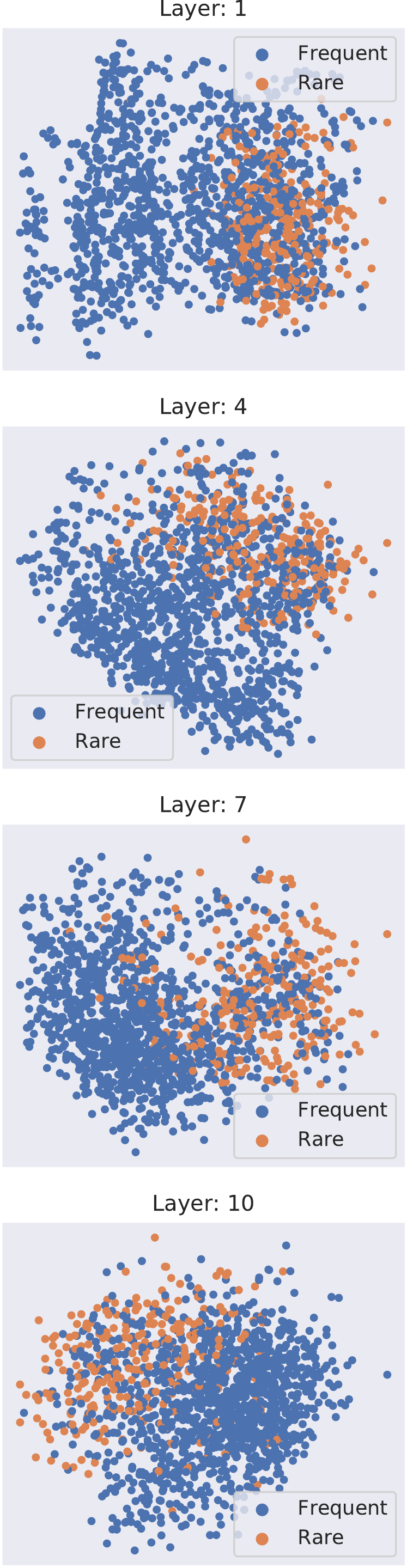}
    \caption{PCA plot of randomly sampled RoBERTa embeddings at layers 1, 4, 7, and 10. Points are colored by token frequency: ``Rare'' means the 20\% least frequent tokens, and ``Frequent'' is the other 80\%.}
    \label{fig:freq-4layer}
\end{figure}

We feed a random selection of BNC sentences into RoBERTa and use PCA to visualize the distribution of rare and frequent tokens at different layers (Figure \ref{fig:freq-4layer}). In all cases, we find that infrequent tokens occupy a different region of the embedding space from frequent tokens, similar to what \citet{gong-frage} observed for static word vectors. This is consistent with the correlation between token-level surprisal and frequency (Figure \ref{fig:freq-correlation}), although the decrease in correlation towards upper layers is not apparent in the PCA plots.

\section{Surprisal gap for BERT and XLNet}
\label{sec:appendix-surprisal-gap}

\begin{figure}
    \centering
    \includegraphics[width=\linewidth]{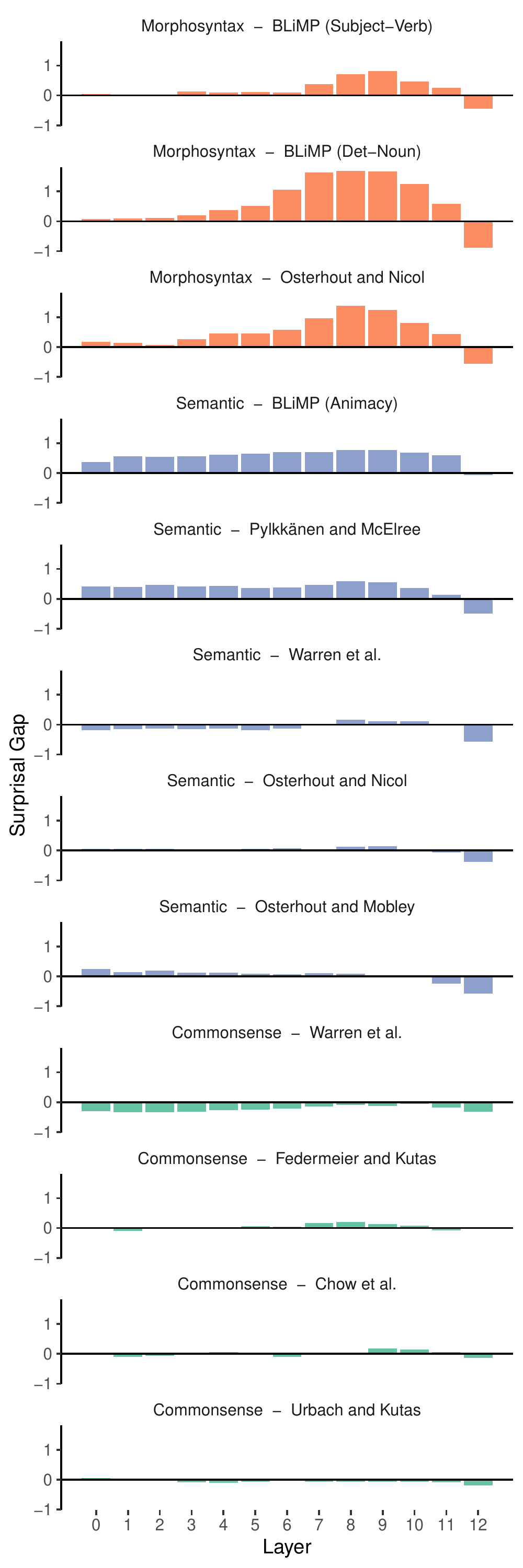}
    \caption{Surprisal gap plot using BERT.}
    \label{fig:bert-manual}
\end{figure}

\begin{figure}
    \centering
    \includegraphics[width=\linewidth]{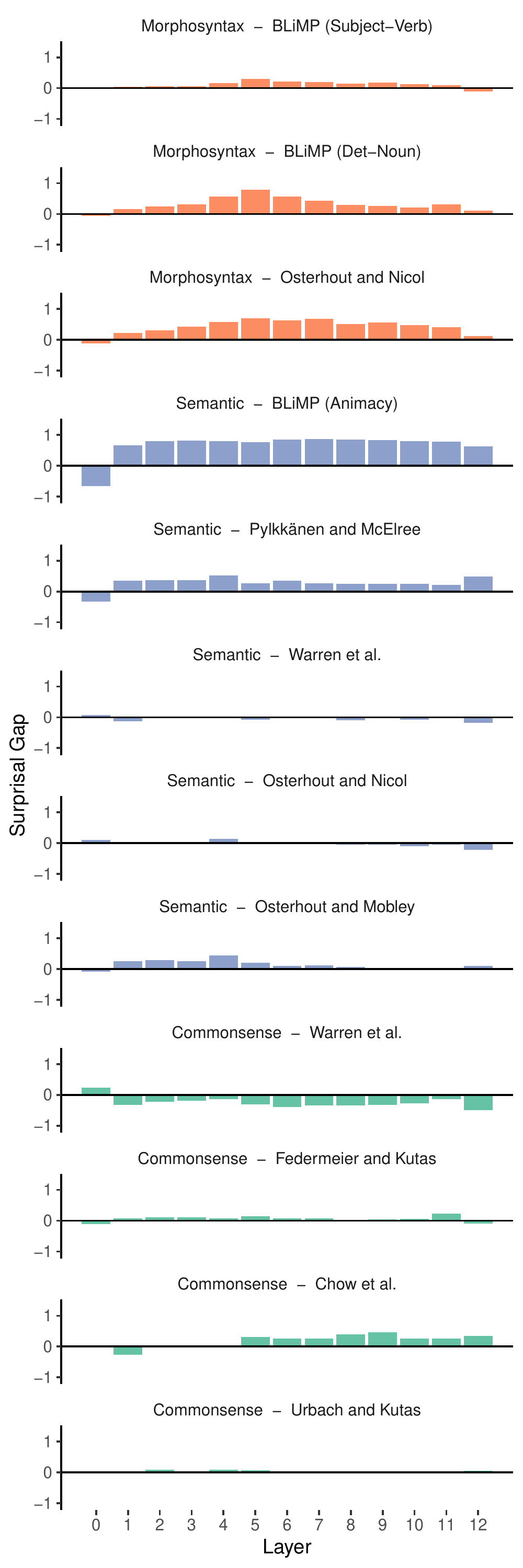}
    \caption{Surprisal gap plot using XLNet.}
    \label{fig:xlnet-manual}
\end{figure}

Figures \ref{fig:bert-manual} and \ref{fig:xlnet-manual} plot the surprisal gaps using the BERT and XLNet models; data and algorithms are identical to the RoBERTa model (Figure \ref{fig:roberta-manual}). The Gaussian model is only sensitive to morphosyntactic anomalies, and not to semantic and commonsense ones.

\end{document}